\documentclass{article}
\usepackage{ijcai22}

\usepackage{times}

\usepackage{soul}
\usepackage{url}
\usepackage[hidelinks]{hyperref}
\usepackage[utf8]{inputenc}
\usepackage[small]{caption}
\usepackage{graphicx}
\usepackage{amsmath}
\usepackage{booktabs}
\usepackage{algorithm}
\usepackage{algorithmic}
\usepackage{comment}
\usepackage{color}

\title{Mean Field inference of CRFs based on GATs}

\author{
LingHongXing$^1$\footnote{luoguangsheng03@126.com}\and
Ma xiangxiang$^{2}$\and
Luo guangsheng$^{2,3}$\\
\affiliations
$^1$Business And Information College, Shanghai, China\\
$^2$Aishu Information Technology Co., Ltd. Shanghai in China\\
$^3$School of Electronic and Electrical Engineering,Shanghai University Of Engineering Science, Shanghai, China\\
\emails
843109540@qq.com,
ma.xiangxiang@aishu.cn,
luoguangsheng03@126.com
}

\begin{document}

\maketitle

\begin{abstract}
In this paper, we propose the mean-field inference algorithm for the Fully Connected Pairwise CRFs model. The improved method Message Passing operation is changed from the original linear convolution to the current graph attention operation, while the process of the inference algorithm is turned into a forward process for the GAT model. Combined with a mean-field inferred label distribution equivalent to the output of aM classifier with only unary potential. We propose a graph attention network model with a residual structure, and the model approach applies to all sequence annotation tasks, such as pixel-level image semantic segmentation tasks as well as text annotation tasks.
\end{abstract}

\section{Introduction}

Conditional random field models have been widely used in text and image sequence annotation tasks \cite{lafferty2001conditional}. However, the time complexity of the exact inference method for the inference of the CRFs model grows exponentially in the length of the sequence, a drawback that greatly limits the application of the model. To address the problem of inference time complexity, the model is sometimes simplified by adding some constraints to the CRFs model. The linear chain CRFs used in the named entity recognition task restricts the existence of edges between only two adjacent sequence elements, which enables the use of a dynamic programming algorithm to obtain the optimal labeled sequence in the model inference phase. For modeling the pixel-level image semantic segmentation task, even when faced with the simplest grid CRFs only approximate inference algorithms can be used.

MCMC \cite{brooks2011handbook} inference is a sampling inference algorithm applicable to all probabilistic graph models, using the local Markov property of conditional random fields, fixing the labels of all token nodes of the current node, and sampling the labels of the current node to update them, commonly used sampling algorithms such as Metroplis sampling \cite{metropolis1953equation,hastings1970monte} and Gibbs sampling \cite{geman1984stochastic}. MCMC inference is a kind of approximate inference algorithm with variance and no deviation, and the variance of the inference result can be reduced by increasing the number of samples, but the increase in the number of samples will inevitably lead to a very slow convergence of the algorithm and cannot be inferred in real time.

Mean-field inference is another very popular probability model approximation inference algorithm \cite{wainwright2008graphical}. Mean-field inference uses a strong mean-field assumption and the optimal distribution inferred from the CRFs model still has a large deviation from the distribution, which is a deviation-free inference algorithm. The mean-field inference time complexity is also exponential for the most common fully connected conditional random fields. The fully connected conditional random field discussed in the remainder of this paper refers specifically to the Fully Connected Pairwise CRFs , since mean-field inference algorithms for this type of conditional random field are able to output results in polynomial-level time complexity.

In the pixel-level semantic segmentation and labeling task of images in fully-connected pairwise CRFs modeled \cite{krahenbuhl2011efficient}, the mean-field inference algorithm of this model can be abstracted into a recurrent neural network \cite{zheng2015conditional}, where each recurrent unit consists of several convolutional operations and the recurrent neural network is expandable into a multilayer weight-sharing network.

Inspired by the above work, in this paper we first treat recurrent units as fully connected graph attention operations, and then unfold the recurrent neural network into a fixed M-layer network with the weight sharing constraint removed, and finally obtain a fully connected graph neural network labeling model equivalent to the fully connected CRF mean field inference.

\begin{figure*}
	\centering
	\includegraphics[width=1\columnwidth]{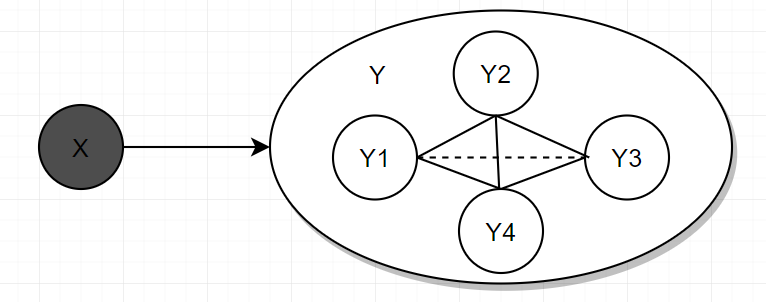}
	\caption{In a fully connected CRFs, the labeled sequence Y constitutes a fully connected Markov random field (any two nodes on the graph are connected by an edge) and there is only one maximal clique in the entire random field. This model can be represented by the expression $P(Y|X) \propto exp \{-E(Y|X)\}$. This is a very common form of the Gibbs distribution in statistical physics, where $E(Y|X)=E(Y_1,Y_2,\dots,Y_N|X)$ is the Gibbs energy. In the Fully Connected Pairwise CRFs model we let $E(Y|X) = \sum_i{\psi_u(Y_i|X)} + \sum_{i<j}{\psi_{p}(Y_i,Y_j|X)}$, this energy expression can be seen as a graph in which all nodes are in an external potential field thus creating a unitary potential, there is no ternary potential in this model, so Fully Connected Pairwise CRFs model cannot fully represent all Fully Connected CRFs model.}
	\label{fig:fully-connected-CRF}
\end{figure*}

\section{RELATED WORK}
\subsection{Fully Connected Pairwise CRFs model}
Consider a random field $Y$ defined over a set of variables $\{Y_1,\dots, Y_N \}$. The domain of each
variable is a set of labels $\mathcal{L} = \{1, 2, \dots , K \}$, the observable variable $X=\{X_1,\dots,X_N\}$ is a known sequence of variables .We develop the following \textbf{Fully Connected CRFs} model model:

\begin{equation}
    P(Y|X)=\frac{1}{Z(X)}exp\{-E(Y|X)\}
\end{equation}

Where $E(Y|X)=\sum_{c\in\mathcal{C_{\mathcal{G}}}}{\psi_{c}(Y_c|X)}$ is The Gibbs energy of a labeling $Y \in \mathcal{L}^N$. For notational convenience we will omit the conditioning $X$ in the rest of the paper, eg: use $E(Y)$ to denote $E(Y|X)$.

In the Fully Connected Pairwise CRFs model, the Gibbs energy expression is:

\begin{equation}
    E(Y)=\sum_i{\psi_u(Y_i)} + \sum_{i<j}{\psi_{p}(Y_i,Y_j)}
\end{equation}

Where i and j range from 1 to N. The unary potential $\psi_u(Y_i)$ is computed independently for each observed element by a classifier that produces a distribution over the label assignment $Y_i$ given $X$:
$$\psi_u(Y_i) = -\log UnaryClassifier(X)$$ 

The pairwise potentials has the following decoupling form:

\begin{equation}
    \psi_p(Y_i,Y_j)=\mu(Y_i,Y_j)k(f_i,f_j)
\end{equation}

where $\mu$ is a label compatibility function that independent of observable variables, $k$ is the kernel function to measure the similarity of the two in the feature space.The feature f is location-informed, For example $f_i=(p_i, X_i)$. in the multi-class image segmentation and labeling task \cite{krahenbuhl2011efficient}, the sequence features and kernel functions take the following form:
\begin{equation}
\label{equation:bilateral_kernel}
    k(f_i,f_j)=\sum_{m=1}^K{\omega^{(m)}exp(-\frac{|p_i-p_j|^2}{2\sigma_{m,1}^{2}}-\frac{|X_i-X_j|^2}{2\sigma_{m,2}^2})}
\end{equation}

\section{Mean field inference of CRFs}

Using model $P(Y)$ to infer the optimal labeling sequence $Y^* = \max\limits_Y {P(Y)}$ is usually a task of very high complexity when the sequence is long or the label takes a large range of values. Instead of computing the exact distribution $P(Y)$,  the mean field approximation computes a distribution $Q(Y)$that minimizes the KL-divergence $D(Q||P)$ among all distributions $Q$ that can be
expressed as a product of independent marginals:

\begin{equation}
\label{equation:Q}
    Q(Y) = \prod_i{Q_i(Y_i)}
\end{equation}

Minimizing the KL-divergence, while constraining $Q(Y)$ and $Q_i(Y_i)$ to be valid distributions, yields the following iterative update equation:

\begin{equation}
\label{equation:MFI}
\begin{aligned}
    &Q_i(Y_i=l)=\frac{1}{Z_i}exp\{-\psi_u(Y_i)\\
    &-\sum_{l^{'}\in \mathcal{L}}{\sum_{j\neq i}{\mu(l,l^{'})k(f_i,f_j)Q_j(Y_j=l^{'})}}\}
\end{aligned}
\end{equation}

A detailed derivation of Equation \ref{equation:MFI} is given in the supplementary material. This update equation
leads to the following inference algorithm:

\begin{algorithm}[H]
    \caption{Mean field in fully connected CRFs}
    \label{alg:alg1}
    \textbf{Input}: observable sequence $X$ \\
    \textbf{Output}: annotated sequence $Y$
    \begin{algorithmic}[1] 
        \STATE initialize $Q$ with $Q\leftarrow UnaryClassifier(X)$
        \STATE initialize $\psi_u$ with $\psi_u^{i,l}\leftarrow -\log Q_{i,l}$
        \WHILE{$Q$ not converged}
                    \STATE $A_{i,l^{'}}\leftarrow \sum_{j \neq i}{k(f_i,f_j) Q_{j,l^{'}}}$ \\
                    \STATE $B_{i,l}\leftarrow \sum_{l^{'} \in \mathcal L}{\mu_{l,l^{'}} A_{i,l^{'}}}$ \\
                    \STATE $Q_{i,l}\leftarrow \frac{1}{Z_i}exp\{-\psi_u^{i,l}-B_{i,l}\}$ \\
        \ENDWHILE
        \STATE $Y_i \leftarrow \max \limits_l {Q_{i,l}}$
        \STATE \textbf{return} Y
    \end{algorithmic} 
\end{algorithm}

The mean-field CRFs inference can be reformulated as a Recurrent Neural Network (RNN), and the loop unit is treated as a convolution operation. So we can construct a CRF-RNN \cite{zheng2015conditional} model by the above algorithm. And the model parameters can be trained end-to-end by using the classification cross-entropy as the loss function.

\begin{figure*}[h]
	\centering
	\includegraphics[width=1.2\columnwidth]{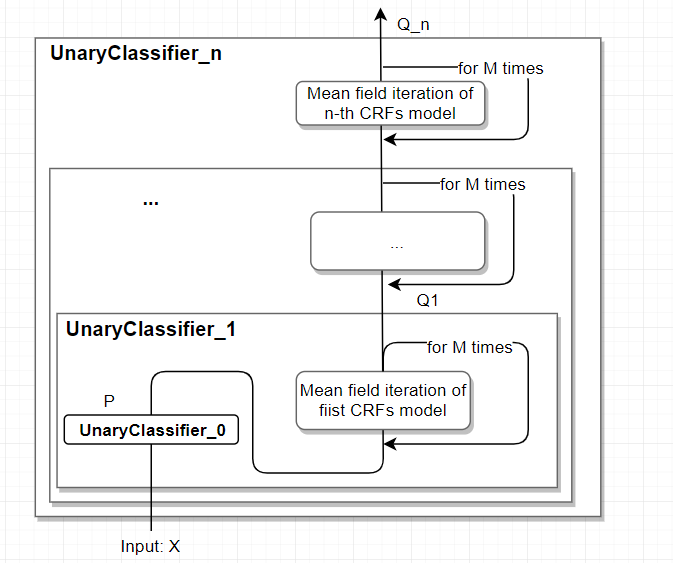}
	\caption{Treating the mean-field inference process as a UnaryClassifier, a new model of CRFs model can be built on top of this new UnaryClassifier. By analogy, multiple CRFs models can be stacked. Let the number of iterations $M = 1$ for the inference of the mean field of the CRFs model, then the model is equivalent to a multilayer graph attention network.}
	\label{fig:fully-connected-CRF}
\end{figure*}

\section{Method}
In this section, we will present our main work on the algorithm for CRFs inference using GATs. Many papers in the past have treated $\sum_{j \neq i}{k(f_i,f_j) Q_{j,l^{'}}}$ as a convolution operation, but we know that the weight parameter of the convolution kernel in the convolution operation is only related to the relative coordinates $pos_i-pos_j$, and it is clear that $k(f_i,f_j)$ is also related to depend on other non-locational features. In fact the operation $\sum_{j \neq i}{k(f_i,f_j) Q_{j,l^{'}}}$ is more like an attention operation. For example, when the kernel function takes the expression\ref{equation:bilateral_kernel}, the corresponding $\sum_{j \neq i}{k(f_i,f_j) Q_{j,l^{'}}}$ operation is actually a non-linear bilateral filtering operation \cite{tomasi1998bilateral}. When the kernel function takes the form of a semi-positive definite symmetric function, referring to the kernel trick used in SVMs \cite{cortes1995support} we know that the physical meaning of expression $k(f_i,f_j)$ is the inner product of the vector $f_i$ and the vector $f_j$ after mapping to a higher dimensional space.
\subsection{Inference using GATs}

 We swap the order of the cumulative operations of the equations \ref{equation:MFI} to obtain a new equation.

\begin{equation}
\label{equation:GPT}
\begin{aligned}
    &Q_i(Y_i=l)=\frac{1}{Z_i}exp\{-\psi_u(Y_i)\\
    &-\sum_{j\neq i}{k(f_i,f_j)\sum_{l^{'}\in \mathcal{L}}{\mu(l,l^{'})Q_j(Y_j=l^{'})}}\}
\end{aligned}
\end{equation}

This update equation leads to the following inference algorithm:

\begin{algorithm}[H]
    \caption{Mean field in fully connected CRFs}
    \label{alg:alg2}
    \textbf{Input}: observable sequence $X$ \\
    \textbf{Output}: annotated sequence $Y$
    \begin{algorithmic}[1] 
        \STATE initialize $Q$ with $Q\leftarrow UnaryClassifier(X)$
        \STATE initialize $\psi_u$ with $\psi_u^{i,l}\leftarrow -\log Q_{i,l}$
        \FOR{$m=1$ \TO $M$}
                    \STATE $A_{i,l}\leftarrow \sum_{l \in \mathcal L}{\mu_{l,l^{'}} Q_{i,l^{'}}}$ \\
                    \STATE $B_{i,l}\leftarrow \sum_{j \neq i}{k(f_i,f_j) A_{j,l}}$ \\
                    \STATE $Q_{i,l}\leftarrow \frac{1}{Z_i}exp\{-\psi_u^{i,l}-B_{i,l}\}$ \\
        \ENDFOR
        \STATE $Y_i \leftarrow \max \limits_l {Q_{i,l}}$
        \STATE \textbf{return} Y
    \end{algorithmic} 
\end{algorithm}

To allow the above algorithm \ref{alg:alg2} to return in a finite time, we only iterate through $M$ steps. We expand these $M$ iterations into a multilayer feedforward neural network with shared parameters in the order of iterations. We analyze this $M-layer$ network using the perspective of graph attention networks:
\begin{itemize}
    \item $A_{i,l}\leftarrow \sum_{l \in \mathcal L}{\mu_{l,l^{'}} Q_{i,l^{'}}}$ is equivalent to the value in the attention mechanism triple (query,key,value).If we also treat the label-compatible matrix $\mu(l,l^{'})$ as a variable positive symmetric matrix $mu(l,l^{'}) = mu(l,l^{'}|X)$. Then V in the attention mechanism triplet also has stronger arbitrariness, and the corresponding graph attention model has stronger fitting ability.And $mu(l,l^{l'}|X)$ can also be represented using neural networks
    \item In expression $\sum_{j \neq i}{k(f_i,f_j) A_{j,l}}$, $f_i$ corresponds to the query of the i-th node, $f_j$ corresponds to the key of the j-th node, and $k(f_i,f_j)$ indicates the correlation degree of the two. The kernel function k can be a Gaussian kernel function, a polynomial kernel function, or a kernel function represented by a neural network.
    \item According to the expression $Q_{i,l}\leftarrow \frac{1}{Z_i}exp\{-\psi_u^{i,l}-B_{i,l}\}$,we know that the activation function of each layer of the graph attention network is $softmax$.
\end{itemize}

\begin{figure*}[h]
	\centering
	\includegraphics[width=2\columnwidth]{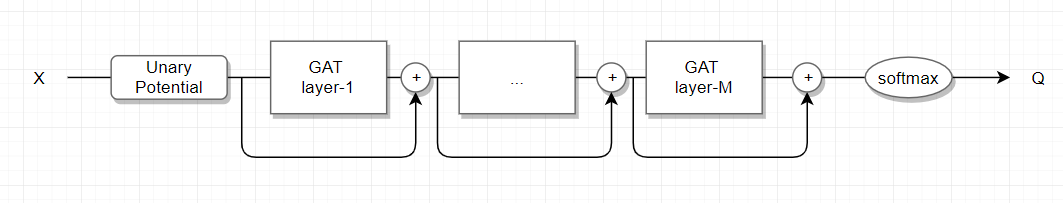}
	\caption{Each cycle of Algorithm \ref{alg:alg3} corresponds to one layer of the graph attention network, and the output of each layer of the graph attention network can be understood as the residuals of the Gibbs energy. Thus a graph attention network model with a residual structure is constructed according to Algorithm \ref{alg:alg3}.}
	\label{fig:residual-gat}
\end{figure*}

\subsection{CRF-GAT models}

 Based on the expression of Q we make the following derivation:

\begin{equation}
\label{equation:Q_}
\begin{aligned}
    &Q(Y) = \prod_i{Q_i(Y_i)} \\
    &= exp\{\sum_i{log Q_i(Y_i)}\} \\
    &=exp\{-\sum_i{\psi_i(Y_i)}\}
\end{aligned}
\end{equation}

According to the expression\ref{equation:Q_} we see that the Gibbs energy  $\sum_i{\psi_i(Y_i)}$ contains only the unary potential of the labels, so we can treat the output of any layer of the M-layer neural network as a unary classifier. We choose the output of the $m_{th}$ layer to update the monadic classifier, so that the remaining $M-m$ layers are treated as a new mean-field iteration of a new conditional random field, and then there is no need for the remaining layers to share parameters with the previous $m$ layers  [figure\ref{fig:fully-connected-CRF}]. Further we treat each layer as a unary classifier, then the parameters of $M$ layers are independent, so that we can completely remove the restriction of sharing parameters of this M-layer neural network, which means that the kernel function $k(f_i,f_j)$ and label-compatible function $\mu(l,l^{'})$ of M-layer neural network are independent. Based on the above discussion, we propose the CRF-GAT models, the forward process algorithm of which is shown in Algorithm \ref{alg:alg3}:

\begin{algorithm}[H]
    \caption{Inference by way of fully connected GATs}
    \label{alg:alg3}
    \textbf{Input}: observable sequence $X$ \\
    \textbf{Output}: annotated sequence $Y$
    \begin{algorithmic}[1] 
        \STATE initialize $P$ with $P\leftarrow UnaryClassifier(X)$
        \STATE initialize $\psi$ with $\psi\leftarrow -\log P$
        \FOR{$m=1$ \TO $M$}
                    \STATE $P_{i,l}\leftarrow \frac{1}{Z_i}exp\{-\psi_{i,l}\}$ \hfill \textcolor{blue}{$\triangleright$compute distribution}
                    \STATE $V_{i,l}\leftarrow \sum_{l^{'} \in \mathcal L}{\mu^{(m)}_{l,l^{'}} P_{i,l^{'}}}$ \hfill\textcolor{blue}{$\triangleright$V in (Q,K,V)}
                    \STATE $\alpha_{i,j}\leftarrow kernel^{(m)}(f_i,f_j)$  \hfill\textcolor{blue}{$\triangleright$attention weight}
                    \STATE $\alpha_{i,i}\leftarrow 0$  \hfill \textcolor{blue}{$\triangleright$no effection on itself}
                    \STATE $R_{i,l}\leftarrow \sum_{j}{\alpha_{i,j}V_{j,l}}$ \hfill \textcolor{blue}{$\triangleright$attention mechanism}
                    \STATE $\psi_{i,l}\leftarrow \psi_{i,l}+R_{i,l}$ \hfill \textcolor{blue}{$\triangleright$update potential}
        \ENDFOR
        \STATE $Y_i \leftarrow \min \limits_l {\psi_{i,l}}$
        \STATE \textbf{return} Y
    \end{algorithmic}
\end{algorithm}

  Based on the algorithm\ref{alg:alg3} we can build a graph attention network model with residual structures, and this model has the following characteristics:
\begin{itemize}
    \item \textbf{End-to-end optimization:} It is natural to observe that the flow of the algorithm \ref{alg:alg3} is essentially a forward inference process for a graph attention network. However, the parameters of each layer of this graph attention network are trained individually using the conditional random field training method. Further, instead of training the parameters of the graph neural network using the general method of training conditional random fields, We can optimize the model parameters by minimizing the cross-entropy loss function with the back-propagation algorithm just like training a normal neural network classifier.At this point, the model is completely transformed into an ordinary graph attention network model.
    
    \item \textbf{Residual network:} If $\psi$ is considered as the output of each layer, the expression $\psi_{i,l}\leftarrow \psi_{i,l}+R_{i,l}$ is actually a residual neural network structure [figure\ref{fig:residual-gat}]. A large number of experiments have demonstrated that neural network models with residual structures can achieve great depth \cite{he2016deep}, so we do not need to worry about the trainability of the models constructed according to the algorithm.
    
    \item \textbf{Attention echanisms} The attention weight between the i-th node and the j-th node in the m-th layer network is $kernel^{(m)}(f_i,f_j)$. This kernel function can be a Gaussian kernel, polynomial kernel, etc., which is commonly used in SVM, or of course, a kernel function represented by a neural network. And the feature $f$ can be either an artificial feature or a feature extracted by a feature engineering network. It is important to note that the attention mechanism we have given so far lacks the attention weight normalization operation used in the transformer model, for which I tried to find a theoretical justification, but this attempt failed.

\end{itemize}

\section{Conclusion}
We consider the mean-field inference algorithm for the fully connected pairwise CRFs model to be consistent with the forward inference process for recursive graph attention networks. One can equate the mean-field inference result of a fully connected pairwise CRF to the forward inference result of a classifier containing only unary potential. The average field inference output of a fully connected pairwise CRFs can be used as the unary potential of another fully connected pairwise CRFs, and repeating the above operation will result in a graph attention network model with residual structure and unnormalized attention weights.

\newpage
\bibliographystyle{named}
\bibliography{ijcai22}

\end{document}